\documentclass[sigconf]{acmart}
\usepackage{CJK}

\AtBeginDocument{%
  \providecommand\BibTeX{{%
    \normalfont B\kern-0.5em{\scshape i\kern-0.25em b}\kern-0.8em\TeX}}}

\setcopyright{acmcopyright}
\copyrightyear{2021}
\acmYear{2021}
\setcopyright{acmcopyright}\acmConference[NLPIR 2021]{2021 5th International Conference on Natural Language Processing and Information Retrieval (NLPIR)}{December 17--20, 2021}{Sanya, China}
\acmBooktitle{2021 5th International Conference on Natural Language Processing and Information Retrieval (NLPIR) (NLPIR 2021), December 17--20, 2021, Sanya, China}
\acmPrice{15.00}
\acmDOI{10.1145/3508230.3508242}
\acmISBN{978-1-4503-8735-4/21/12}

\begin{document}

%%
%% The "title" command has an optional parameter,
%% allowing the author to define a "short title" to be used in page headers.
\title{Low-Resource NMT: A Case Study on the Written and Spoken Languages in Hong Kong}

%%
%% The "author" command and its associated commands are used to define
%% the authors and their affiliations.
%% Of note is the shared affiliation of the first two authors, and the
%% "authornote" and "authornotemark" commands
%% used to denote shared contribution to the research.
\author{Hei Yi Mak}
\affiliation{%
  \institution{The Chinese University of Hong Kong}
  \country{Hong Kong}
}

\author{Tan Lee}
\affiliation{%
  \institution{The Chinese University of Hong Kong}
  \country{Hong Kong}
}

%%
%% By default, the full list of authors will be used in the page
%% headers. Often, this list is too long, and will overlap
%% other information printed in the page headers. This command allows
%% the author to define a more concise list
%% of authors' names for this purpose.
\renewcommand{\shortauthors}{Mak and Lee}
\newcommand{\quotes}[1]{``#1''}
\newcommand{\chinese}[1]{\begin{CJK*}{UTF8}{bsmi}#1\end{CJK*}}
\newcommand{\quotedchinese}[1]{\quotes{\chinese{#1}}}

%%
%% The abstract is a short summary of the work to be presented in the
%% article.
\begin{abstract}
  The majority of inhabitants in Hong Kong are able to read and write in standard Chinese but use Cantonese as the primary spoken language in daily life. Spoken Cantonese can be transcribed into Chinese characters, which constitute the so-called written Cantonese. Written Cantonese exhibits significant lexical and grammatical differences from standard written Chinese. The rise of written Cantonese is increasingly evident in the cyber world. The growing interaction between Mandarin speakers and Cantonese speakers is leading to a clear demand for automatic translation between Chinese and Cantonese. This paper describes a transformer-based neural machine translation (NMT) system for written-Chinese-to-written-Cantonese translation. Given that parallel text data of Chinese and Cantonese are extremely scarce, a major focus of this study is on the effort of preparing good amount of training data for NMT. In addition to collecting 28K parallel sentences from previous linguistic studies and scattered internet resources, we devise an effective approach to obtaining 72K parallel sentences by automatically extracting pairs of semantically similar sentences from parallel articles on Chinese Wikipedia and Cantonese Wikipedia. We show that leveraging highly similar sentence pairs mined from Wikipedia improves translation performance in all test sets. Our system outperforms Baidu Fanyi's Chinese-to-Cantonese translation on 6 out of 8 test sets in BLEU scores. Translation examples reveal that our system is able to capture important linguistic transformations between standard Chinese and spoken Cantonese.
\end{abstract}

% \begin{CJK*}{UTF8}{gbsn} 文章内容o既o甘嘅們们。\end{CJK*}

%%
%% The code below is generated by the tool at http://dl.acm.org/ccs.cfm.
%% Please copy and paste the code instead of the example below.
%%


%%
%% Keywords. The author(s) should pick words that accurately describe
%% the work being presented. Separate the keywords with commas.

%% A "teaser" image appears between the author and affiliation
%% information and the body of the document, and typically spans the
%% page.

%%
%% This command processes the author and affiliation and title
%% information and builds the first part of the formatted document.
\maketitle

\section{Introduction}
Machine translation refers to the practice of translating text or speech from one language to another language using computer software. Traditionally, machine translation systems were designed with rule-based or statistical approaches. Recent research interests have been largely drawn towards the use of artificial neural network, leading to a branch of MT research known as neural machine translation (NMT). State-of-the-art NMT models can be divided into three categories: RNN-based models \cite{Sutskever14}, CNN-based models \cite{gehring2017convolutional}, and transformers \cite{vaswani2017attention}. These models generally follow the encoder-decoder architecture with attention mechanism \cite{bahdanau2014neural}. The encoder-decoder architecture allows end-to-end training with source and target sentences. The attention mechanism enables the decoder to learn to focus on the specific parts of input sentence.

This paper presents an NMT system for written-Chinese-to-written-Cantonese translation. Written Chinese corresponds to the written form of the standard spoken Chinese, Mandarin. Cantonese, or referred to as the Yue dialect, is one of the major Chinese dialects. It is spoken by tens of millions of people in Hong Kong, Macau, Guangdong and Guangxi provinces, as well as many overseas Chinese communities in Southeast Asia and North America. Despite being used primarily in spoken form, Cantonese can be transcribed into Chinese char-acters. Such transcription is referred to as written Cantonese. Spoken Cantonese and Mandarin are considered to be mutually unintelligible \cite{snow2004cantonese}. Their written form, i.e., written Cantonese and written Chinese also exhibit significant lexical and grammatical differences. In Hong Kong, the majority of the population are native Cantonese speakers and communicate in Cantonese in both informal and formal speaking contexts (e.g., daily communication, council meetings, business meeting, news reports, etc.). In typical written contexts, however, Chinese is the preferred language. Nevertheless, writing in Cantonese has become very popular among internet and mobile users. With the prevalence of instant messaging and social media, as well as millions of native speakers around the globe, written Cantonese has found its importance and usage in online social communication. For simplicity of terminology, written Chinese is referred to as \quotes{Chinese} and written Cantonese is referred to as \quotes{Cantonese} in this paper.

Chinese-to-Cantonese MT systems have many potential applications. They allow automatic translation of Chinese documents into Cantonese, which could be readily combined with Cantonese text-to-speech systems to realize applications like storytelling, virtual news anchor, etc. By training in the reverse direction with similar parallel data, a Cantonese-to-Chinese MT system can also be developed. Coupled with Cantonese speech-to-text, the MT system can support the generation of Chinese text caption for Cantonese-speaking recording. Such systems can be very helpful to Mandarin speakers in consuming Cantonese audiovisual materials, e.g., lectures, public speech.

NMT models are typically trained on large datasets that contain millions of parallel sentences of the intended language pair. When such parallel data resources are limited, researchers are required to work in the so-called low-resource setting. Parallel data of Chinese and Cantonese are extremely low-resourced. This is mainly because spoken Cantonese is rarely transformed into digital text for archive. Even when a Cantonese corpus is available, substantial efforts are needed to translate the data into Chinese. The scarcity of parallel resources is a common issue in dialect translation \cite{honnet2017machine, abe2018multi}. One of the contributions of this paper is to provide a detailed account of how training data are collected in the case of Chinese dialect translation. In particular, we constructed a set of 72K parallel sentences by mining pairs of semantically similar sentences from parallel articles on Chinese Wikipedia and Cantonese Wikipedia using LASER \cite{artetxe2019massively}. We hope to provide useful insights to the MT community on acquiring parallel resources in Chinese and Cantonese and possibly other language pairs.

This paper is organized as follows. In the next section, we review the related work on Chinese and Cantonese MT and a similar low-resource task on German dialects. In Section 3, we describe the main linguistic considerations in Chinese-to-Cantonese translation. In Section 4, the NMT model used for Chinese-to-Cantonese MT is described. In Section 5, we report the effort of searching and collecting data resources for MT. In Section 6, the method of mining parallel sentences from Wikipedia is presented. In Section 7, we describe the training process. In Section 8, we evaluate the developed MT system by comparing BLEU scores against a baseline system and Baidu Fanyi\footnote{https://fanyi.baidu.com/\#cht/yue}, and analyze its performance on linguistic transformations with examples. Conclusions are given in Section 9.

\section{Related Work}

Xu and Fung \cite{xu2012cross} developed an MT system that could generate Chinese transcription from Cantonese speech. The system leveraged the language model statistics of resource-rich language using a syntactic reordering model to improve speech recognition of the low-resource language. Wong and Lee \cite{wong2018register} investigated a rule-based approach to translating Chinese sentences to Cantonese sentences at different register levels. This system incorporates annotated lexical mappings and syntactic transformations to perform register-sensitive translation.

Honnet et al. \cite{honnet2017machine} designed an MT system from Swiss German to High German with fewer than 60K parallel sentences. Swiss German represents a collection of spoken dialects in Switzerland that are seldom written and generally unintelligible to German speakers outside Switzerland. However, Swiss German speakers write in High German, the standard German language. The relation of Swiss German and High German is akin to that of Cantonese and Chinese. In the work of Honnet et al., the parallel resources were collected from books, Wikipedia, transcriptions, the Bible and dictionaries. 

\section{Linguistic Considerations in Chinese-to-Cantonese Translation}

\subsection{Lexical Changes}
One important linguistic consideration in developing a Chinese-to-Cantonese translation system is its capability of handling lexical mappings between written Chinese and spoken Cantonese. Lexical difference is seen in almost all types of words in the two languages. Common examples include mapping particles \quotedchinese{的} to \quotedchinese{o既} (of) and \quotedchinese{地} to \quotedchinese{咁} (-ly)), adverbs \quotedchinese{了} to \quotedchinese{o左} (already) and \quotedchinese{不} to \quotedchinese{唔} (not), pronouns \quotedchinese{他} to \quotedchinese{佢} (he) and \quotedchinese{他們} to \quotedchinese{佢地} (they), verbs \quotedchinese{是} to \quotedchinese{係} (is) and \quotedchinese{喜
歡} to \quotedchinese{鍾意} (like), adjectives \quotedchinese{美麗} to \quotedchinese{靚} (beautiful) and \quotedchinese{冷} to \quotedchinese{凍} (cold), nouns \quotedchinese{小孩子} to \quotedchinese{細路仔} (child) and \quotedchinese{椅子} to \quotedchinese{凳} (chair). Since lexical differences take place in many frequently used words, correct mapping between the lexical items in Chinese and Cantonese is critical to high-quality translation between the two languages.

\subsection{Word Reordering}

Word order is another important aspect in which Chinese and Cantonese exhibit notable differences. Table~\ref{tab:wordorder} gives a few examples of parallel Chinese and Cantonese sentences that involve word reordering.

\begin{table}
  \caption{Examples of word reordering in Chinese and Cantonese.}
  \label{tab:wordorder}
  \begin{tabular}{ll}
    \toprule
    Example 1 & \\
    \midrule
    English & Give me a cup of water.\\
    Chinese & \chinese{給我一杯水} (\quotes{give} \quotes{me} \quotes{a} \quotes{cup} \quotes{water})\\ 
    Cantonese & \chinese{畀杯水我} (\quotes{give} \quotes{cup} \quotes{water} \quotes{me})\\ 
    \toprule
    Example 2 & \\
    \midrule
    English & I am taller than you.\\
    Chinese & \chinese{我比你高} (\quotes{I} \quotes{compare} \quotes{to} \quotes{you} \quotes{tall})\\ 
    Cantonese & \chinese{我高過你 } (\quotes{I} \quotes{tall} \quotes{than} \quotes{you})\\ 
    \toprule
    Example 3 & \\
    \midrule
    English & Eat more.\\
    Chinese & \chinese{多吃點} (\quotes{more} \quotes{eat} \quotes{some})\\ 
    Cantonese & \chinese{食多o的} (\quotes{eat} \quotes{more} \quotes{some})\\ 
  \bottomrule
\end{tabular}
\end{table}

The translation in Example 1 involves an expression \quotes{give (something) to (some-body)}.  In Chinese, the word order is, transitive verb \quotedchinese{給} $\rightarrow$ indirect object \quotedchinese{我} $\rightarrow$ direct object \quotedchinese{一杯水}.  In Cantonese, the indirect object is placed after the direct object, giving the word order, transitive verb \quotedchinese{畀} $\rightarrow$ direct object \quotedchinese{杯水} $\rightarrow$ indirect object \quotedchinese{我}.
Word reordering is also needed if the source sentence describes a comparative relation. In Example 2, the Chinese sentence follows the order, subject \quotedchinese{我} $\rightarrow$ particle \quotedchinese{比} $\rightarrow$ object \quotedchinese{你} $\rightarrow$ adjective \quotedchinese{高}, but in Cantonese the order is changed to, subject \quotedchinese{我} $\rightarrow$ adjective \quotedchinese{高} $\rightarrow$ particle \quotedchinese{過} $\rightarrow$ object \quotedchinese{你}.
Certain adverbs are placed differently in relation to the neighboring verb in the two languages. In Example 3, the adverb \quotedchinese{多} precedes the verb in Chinese but succeeds the verb in Cantonese.

\subsection{Code-switching in Cantonese}
Code-switching is a common practice among Cantonese speakers \cite{gibbons1987code} in Hong Kong and overseas. It refers to the practice of mixing English words into Cantonese speech. Thus, the translation task between Chinese and Cantonese involves not only Chinese characters, but English words as well. For example, the Chinese sentence \quotedchinese{我稍後才有時間打電話給他}, can be translated into the Cantonese sentence, \quotedchinese{我一陣先有時間call佢}, which contains the English word \quotes{call}.

To capture this linguistic phenomenon of the Cantonese-speaking community, the Chinese-to-Cantonese NMT model needs to be able to output sentences that contain both Chinese characters and English words.

\section{The NMT Model}

\subsection{Pretrained BERT Encoder Decoder Model}

We use the Encoder-Decoder Model from the HuggingFace Transformers library\footnote{https://github.com/huggingface/transformers}. The encoder is a BERT \cite{devlin2018bert} model (\quotes{bert-base-chinese}) pre-trained on the Chinese Wikipedia corpus. The decoder is the same BERT model with a language modeling head added on top to output predicted scores for each vocabulary token. During training, the source sentence and target sentence are fed into the model, which will learn to predict the next token based on the encoder output and previous tokens in the target sentence. During translation, the encoder output embedding and all previously generated tokens are fed into the model to predict the next token regressively until the end-of-sentence token is generated. The maximum length of the input sequence is 512 characters.

\subsection{Tokenizer and Vocabulary}

The tokenizer of the pretrained \quotes{bert-base-chinese} model converts a sentence into a sequence of input tokens. The model can only read and output tokens that are stored in the tokenizer's vocabulary. The vocabulary, with a fixed size of 21128, consists of mostly Chinese characters, some English subwords, numbers, special characters (e.g., \quotes{[PAD]}, \quotes{[UNK]}), unused tokens (e.g., \quotes{[unused1]}), emoji, punctuations and symbols. The model is particularly suitable for our translation task because the inclusion of both Chinese and English tokens makes code-switching possible.

\section{Training Data}

As discussed earlier, parallel text data in Chinese and Cantonese are extremely low-resourced. This is largely because the transcription of spoken Cantonese into digital text and the translation between Chinese and Cantonese both require great efforts. In this study, parallel data are collected from previous linguistic research and the internet. The largest dataset we were able to locate contains barely more than 10K parallel sentences. Combining all the datasets we found, there are around 28K sentence pairs. To overcome the low-resource barrier, we investigated a method of mining parallel sentences from the Chinese Wikipedia and Cantonese Wikipedia and constructed the Wikipedia Dataset which contains 72K parallel sentences. As a result, a total of 101434 parallel sentences are available for training our NMT system. A summary of the data resources is given as in Table \ref{tab:trainingdata}.

\begin{table*}
  \caption{A summary of the parallel data in Chinese and Cantonese.}
  \label{tab:trainingdata}
  \begin{tabular}{llllll}
    \toprule
    Dataset	&Type of Materials&	Train& Validation& Test& Total \\
    \midrule
    Kaifang Cidian & Conversations & 8658 & 1082 & 1083 & 10823\\
    Tatoeba & Conversations & 2025 & 253 & 254 & 2532\\
    Lee's Parallel Corpus & TV dramas, news, broadcast & 6580 & 822 & 823 & 8225\\
    Cantonese-Mandarin Parallel Dependency Treebank & Short films, 
    council meetings & 803 & 100 & 101 & 1004\\
    Storytelling Dataset & Stories & 4048 & 506 & 506 & 5060\\
    Hong Kong Storybooks & Stories & 520 & 65 & 66 & 651\\
    ILC Cantonese Program & Narrative essays & 356 & 45 & 45 & 446\\
    Wikipedia Dataset & Wikipedia & 69693 & 1500 & 1500 & 72693\\
    \midrule
     &  & 92683 & 2873 (non-wiki) & 4378 & 101434\\
     &  & & 1500 (wiki) & & \\
  \bottomrule
\end{tabular}
\end{table*}

\textbf{Lee's Parallel Corpus}. Lee \cite{lee2011toward} created a parallel corpus of Cantonese transcriptions and Chinese subtitles of news reports, dramas and broadcast on a television channel in Hong Kong. The corpus contains 8225 pairs of sentences.

\textbf{Cantonese-Mandarin Parallel Dependency Treebank}. Wong et al. \cite{wong2017quantitative} constructed a parallel dependency treebank of Cantonese transcription and Chinese subtitles of short films and legislative council meetings, with 1004 pairs of sentences. The treebank is freely available on GitHub\footnote{https://github.com/UniversalDependencies/UD\_Cantonese-HK} \footnote{https://github.com/UniversalDependencies/UD\_Chinese-HK}.

\textbf{Storytelling Dataset}. The storytelling dataset is a set of parallel Chinese scripts and transcriptions of Cantonese storytelling recordings, which were intended for the development of a personalized storytelling system. The 28 stories in the recordings were narrated by a native Cantonese speaker. The Chinese scripts were significantly altered in the Cantonese narration for expressive storytelling. We manually aligned the sentences and filtered the problematic ones, resulting in 5060 pairs of sentences.

\textbf{Kaifang Cidian}. The website\footnote{http://kaifangcidian.com/} of Kaifang Cidian (translated as \quotes{Open Dictionary} in English) has a Cantonese-Chinese sentence dataset with 10823 pairs of parallel sentences. The dataset features Cantonese sentences that are used in conversations and informal settings. Code-switching is found in some of the Cantonese sentences.

\textbf{Tatoeba}. The website\footnote{https://tatoeba.org/eng/} of Tatoeba has a collection of sentences and translations into over 380 languages contributed by a voluntary community. We downloaded all the Cantonese sentences, Chinese sentences, and a table containing the links between the sentences. 2532 pairs of parallel sentences were extracted.

\textbf{Hong Kong Storybooks}. The website\footnote{https://global-asp.github.io/storybooks-hongkong/} of Hong Kong Storybooks has a free collection of stories available in English, Chinese and Cantonese. There are 40 stories at different levels. We extracted the Chinese and Cantonese content and aligned the sentences, resulting in 651 pairs of parallel sentences.

\textbf{ILC Cantonese Program}. The Cantonese Online Program\footnote{https://www.ilc.cuhk.edu.hk/Chinese/pthprog1/tm\_courses.html} is a website hosted by the Independent Learning Center of the Chinese University of Hong Kong. The purpose of the program is to help native Mandarin speakers to learn Cantonese. The program consists of 15 chapters, each of which contains a short narrative essay in Cantonese and its translation in Chinese. We sentence-aligned the essays and collected 446 pairs of parallel sentences.

\textbf{Wikipedia Dataset}. A dataset of 72693 pairs of sentences constructed by automatically extracting semantically similar sentences from articles on Chinese Wikipedia and Cantonese Wikipedia of the same topic. Details of the mining process are described in Section 6.

\section{Mining Parallel Sentences from the Wikipedia}

\subsection{Parallel Resources in the Wikipedia}
Wikipedia is one of the largest multilingual text resources on the internet. On the same topic, there may exist parallel articles in different languages. These parallel articles are rarely mutual translations of each other. Nevertheless, due to the similarity of contents, these articles may contain sentences that are equivalent in meaning, which can be exploited as parallel data for training NMT systems. Among versions of Wikipedia in different Chinese dialects, the standard Chinese (zh) Wikipedia and Cantonese (zh-yue) Wikipedia are the most prominent versions with around 1.1M articles and around 100K articles respectively. We explore the possibility of extracting parallel sentences from these two versions of Wikipedia.

\subsection{Mining Parallel Sentences from Wikipedia}

One of the most notable methods for mining parallel sentences from the Wikipedia is the WikiMatrix \cite{schwenk2019wikimatrix}. It is an approach to mining parallel sentences in a massively multilingual setting using the multilingual sentence embedding toolkit LASER \cite{artetxe2019massively}. LASER uses a single language agnostic encoder to compute multilingual embeddings for sentences in 93 languages. Simply speaking, LASER maps sentences from different languages to vectors in the same embedding space so that the semantic similarity between sentences in different languages can be measured. The extracted corpus, Wiki-Matrix bitexts, consists of parallel sentences for 1620 different language pairs mined from Wikipedia articles in 85 languages. Although LASER supports evaluation on similarity between Chinese and Cantonese sentences, parallel data of the Chinese-Cantonese language pair were not available in the WikiMatrix bitexts\footnote{https://github.com/facebookresearch/LASER/tree/master/tasks/WikiMatrix}.

We see the opportunity to construct a parallel training dataset by applying LASER to automatically extract parallel sentences from the Chinese Wikipedia and Cantonese Wikipedia.

\subsection{Corpus Preparation}
We downloaded a complete copy of Chinese Wikipedia articles, a complete copy of Cantonese Wikipedia articles, and the inter-language links from the Wikimedia Database backup dumps\footnote{https://dumps.wikimedia.org/zhwiki/, } \footnote{https://dumps.wikimedia.org/zh\_yuewiki/}. The corpora were cleaned with the WikiExtractor\footnote{https://github.com/attardi/wikiextractor}, removing unwanted metadata and html tags. With the inter-language links, we are able to match an article in the Cantonese Wikipedia corpus with the corresponding article in the Chinese Wikipedia corpus. After the mapping, around 70K pairs of articles were obtained.

\subsection{Parallel Sentences Extraction}
For each pair of Chinese and Cantonese articles, we split the text into two lists of sentences and compute their sentence embeddings with LASER. The embedding of each sentence in the Chinese article is compared with that of each sentence in the Cantonese article. The comparison is done by calculating the cosine similarity of two sentence embeddings (a score between 0 and 1), which is referred to as the similarity score. We select all pairs of sentences with similarity scores higher than a predetermined threshold (to be discussed in Section 6.5). As both Chinese and Cantonese are in the form of sequence of Chinese characters, some selected sentences pairs are identical. These pairs are removed to avoid constructing training data that have identical source and target sentences. After removing pairs of identical sentences, if there are multiple sentence pairs with the same Chinese sentence, we retain only the pair with the highest similarity score.

\subsection{Similarity Score Threshold}
We consider different thresholds of similarity scores and perform human inspection on samples of sentence pairs with similarity scores higher than 0.90, 0.93 and 0.95 respectively. Table \ref{tab:similarity} shows the number of sentence pairs we obtained for different thresholds.

\begin{table}
  \caption{Number of sentence pairs obtained using different similarity score thresholds.}
  \label{tab:similarity}
  \begin{tabular}{ll}
    \toprule
    Similarity score & Number of sentence pairs \\
    \midrule
    $\geq$ 0.90 & 122264\\
    $\geq$ 0.93 & 73252\\ 
    $\geq$ 0.95 & 45475\\ 
  \bottomrule
\end{tabular}
\end{table}

Sentence pairs of similarity score higher than 0.90 contain problematic pairs that involve numbers. For instance, expressions of different years (\quotedchinese{1840年}, \quotedchinese{1842年}) and expressions of different dates (\quotedchinese{2017年2月27日}, \quotedchinese{2017年2月3號}) were not filtered out. Common to these sentence pairs is that the pair share similar structure but are inconsistent in meaning because certain digit(s) has or have different numerical values. It would be undesirable to include such pairs in the training corpus. Samples of similarity score higher than 0.93 and those higher than 0.95 are high quality parallel sentences without the aforementioned problem. Since setting a lower similarity score yields more sentence pairs, 0.93 is chosen to be the similarity score threshold for parallel sentence extraction.

\section{Training}
With a similarity threshold of 0.93, 72693 parallel sentences were mined from the Chinese Wikipedia and Cantonese Wikipedia. The Wikipedia dataset is split into a training set of 69693 sentence pairs, a validation set of 1500 pairs and a test set of 1500 pairs. Each of the 7 other non-wiki datasets is split into training, validation and test sets in the proportion 8:1:1. The data partition arrangement is detailed as shown in Table \ref{tab:trainingdata}. A non-wiki validation set is formed by merging the 7 validation sets with the wiki validation set being excluded. The 8 test sets are used separately for evaluation.

During training, validation is carried out every 10 epochs using the BLEU score \cite{papineni2002bleu} on the non-wiki and the wiki validation set. Only the BLEU score on the non-wiki validation set is used to determine when to stop training. The BLEU score on the wiki validation set is measured only for reference because it is significantly higher than the BLEU score on other datasets due to the high similarity of the source and target sentences extracted with a high similarity threshold. The system is trained until the BLEU score on the non-wiki validation set converges at around one thousand epochs.

\section{Performance Evaluation}
\subsection{BLEU Score}
After training, the BLEU scores of our system's translations on the 8 test sets are evaluated. A baseline translation system and the Baidu Fanyi's Chinese-to-Cantonese translation  are used for performance comparison. The baseline translation is essentially a copy-and-paste model whose output is simply the Chinese source sentence. We include the BLEU score of the baseline translation as a reference because the Chinese source sentence and Cantonese target sentence are both written in Chinese characters and share some common lexical items. The source sentences in the test sets are also translated with Baidu Fanyi and the BLEU scores are computed. Both our system and Baidu Fanyi outperform the baseline translation significantly. Our system demonstrates competitive performance to Baidu Fanyi. It outscores Baidu Fanyi in 6 out of 8 test sets. The results are shown as in Table \ref{tab:bleu}.

\begin{table*}
  \caption{BLEU scores on the 8 test sets.}
  \label{tab:bleu}
  \begin{tabular}{lllll}
    \toprule
     & \multicolumn{4}{c}{BLEU score} \\\cline{2-5}
    Test set & Baseline & Non-wiki & Our system & Baidu Fanyi\\
    \midrule
    Kaifang Cidian & 11.64 & 42.93 & 43.81 & \textbf{47.00} \\
    Tatoeba & 21.79 & 52.77 & 53.09 & \textbf{64.05} \\
    Lee's Parallel Corpus & 36.03 & 53.22 & \textbf{58.43} & 45.35 \\
    Cantonese-Mandarin Parallel Dependency Treebank & 17.13 & 28.15 & \textbf{32.14} & 28.40 \\
    Storytelling Dataset & 28.37 & 52.64 & \textbf{53.80} & 40.13 \\
    Hong Kong Storybooks & 19.08 & 50.74 & \textbf{51.99} & 38.81 \\
    ILC Cantonese Program & 35.23 & 57.47 & \textbf{63.38} & 53.65 \\
    Wikipedia Dataset & 63.18 & 43.99 & \textbf{81.70} & 63.11 \\
  \bottomrule
\end{tabular}
\end{table*}

In addition, we trained another system using only the non-wiki data and evaluated its performance on the same test data. Compared to this non-wiki system, our system, which is trained on both wiki and non-wiki data, shows a significant improvement in BLEU score, from 43.99\% to 81.70\%, on the wiki test data. The discrepancy is expected as wiki data are included to train our system. Nevertheless, such inclusion of training data mined from wiki also increases BLEU scores on all other test datasets. The most noticeable improvements are on Lee's Parallel Corpus, ILC Cantonese Program, and Cantonese-Mandarin Parallel Dependency Treebank.

\subsection{Performance Analysis with Examples}
In this section, a few typical examples of translation results are analyzed. For each example, our system's translation is compared to that from Baidu Fanyi in terms of the use of desired linguistic transformations, e.g., lexical mapping, word reordering and code-switching, as discussed in Section 3. The examples are given as in Table \ref{tab:examples}.

\textbf{Example 1.} Baidu Fanyi fails to capture the word reordering of the phrase \quotedchinese{多收點} (\quotes{more} \quotes{receive} \quotes{some}) into \quotedchinese{收多o的} (\quotes{receive} \quotes{more} \quotes{some}). While it is acceptable to make no lexical change on the word \quotedchinese{消息} in Baidu Fanyi's translation, our system applies a more colloquial lexical transformation (\quotedchinese{消息} to \quotedchinese{料}), which is used in the target sentence.

\textbf{Example 2}. Both translations are able to make the correct lexical change to map the word \quotedchinese{便宜} to \quotedchinese{平}. The word \quotedchinese{服裝} is acceptable in Cantonese but \quotedchinese{衫} is more commonly used. It was incorporated in Baidu Fanyi's translation but our system did not capture it. The phrase \quotedchinese{比大陸便宜} (\quotes{compare to} \quotes{mainland} \quotes{cheap}) should be reordered as \quotedchinese{平過大陸} (\quotes{cheap} \quotes{than} \quotes{mainland}) in Cantonese. Our system performs this transformation correctly but Baidu Fanyi's translation keeps the Chinese word order.

Example 3. In both translations, the verb  \quotedchinese{躺} is transformed to \quotedchinese{攤} (lay) and \quotedchinese{讓} transformed to \quotedchinese{等} (let). The particle \quotedchinese{o左} (already) after \quotedchinese{攤} (lay) in our system's translation deviates slightly from the meaning of the source sentence. Our system impose the correct word reordering of the phrase \quotedchinese{躺到床上去}, i.e., swapping the position of the preposition \quotedchinese{上} (on) and the noun \quotedchinese{床} (bed). Whilst the Baidu Fanyi's translation does not make this change. The change from the verb phrase \quotedchinese{檢查一下} (take a look) to \quotedchinese{check 一 check} in our system's translation is a code-switched term commonly used by native Cantonese speakers.

Example 4. The term \quotedchinese{這陣子} (recently) is correctly mapped to \quotedchinese{呢輪} by our system and to \quotedchinese{呢排} by Baidu Fanyi. Both translations demonstrate the ability of incorporating code-switching lexical transformation by changing the term \quotedchinese{加班} (work overtime) to \quotedchinese{開o.t.} or \quotedchinese{開OT}. This is indeed a common practice of code-switching in Cantonese.

In these examples, our system demonstrates the ability to perform appropriate lexical transformation and code-switching, and superior performance in word reordering.

\section{Conclusion}
Lacking parallel data resources is the major challenge in developing robust neural machine translation systems that involve regional spoken language like Cantonese. In this research we have demonstrated the feasibility of exploiting monolingual data, which are relatively large in amount, by mining similar sentences from multilingual online data resources, i.e., Wikipedia. Trained on the mined parallel data in addition to those collected directly from published materials, our Chinese-to-Cantonese MT system shows BLEU score improvements in all test sets. Our system attains competitive performance to the Chinese-to-Cantonese translation on Baidu Fanyi. Examples of translation results reveal that crucial linguistic transformations like lexical mapping, word reordering and code-switching are captured by our system.

The proposed approach of data collection is applicable to other tasks of natural language processing that require large amount of Chinese and Cantonese parallel text. The methods used in this study could provide important reference for designing MT systems on other language pairs, for which parallel resources are scarce but monolingual data are abundant.

\begin{table*}
  \caption{Examples of word reordering in Chinese and Cantonese.}
  \label{tab:examples}
  \begin{tabular}{ll}
    \toprule
    Example 1 & (Lee's Parallel Corpus) \\
    \midrule
    Source sentence & \chinese{我 知道，多 收 點 消息 看 清楚 前 路}\\
    & \quotes{I} \quotes{know}, \quotes{more} \quotes{receive} \quotes{some} \quotes{information} \quotes{see} \quotes{clear} \quotes{forward} \quotes{path}\\ 
    Target sentence & \chinese{我 知，收 多 o的 料 睇 清 條 路}\\
    & \quotes{I} \quotes{know}, \quotes{receive} \quotes{more} \quotes{some} \quotes{information} \quotes{see} \quotes{clear} \quotes{the} \quotes{path}\\
    Our system & \chinese{我 知， 收 多 o的 料 睇 清楚 前 路}\\ 
    & \quotes{I} \quotes{know}, \quotes{receive} \quotes{more} \quotes{some} \quotes{information} \quotes{see} \quotes{clear} \quotes{forward} \quotes{path}\\
    Baidu Fanyi & \chinese{我 知，多 收 o的 消息 睇 清楚 前 路}\\
    & \quotes{I} \quotes{know}, \quotes{more} \quotes{receive} \quotes{some} \quotes{information} \quotes{see} \quotes{clear} \quotes{forward} \quotes{path}\\
    \midrule
    
    \toprule
    Example 2 & (ILC Cantonese Program) \\
    \midrule
    Source sentence & \chinese{其他的 電器 服裝 等 有 許多 都 比 大陸 便宜}\\
    & \quotes{other} \quotes{electrical appliances} \quotes{clothes} \quotes{etc.} \quotes{have} \quotes{many} \quotes{all} \quotes{compare to} \quotes{mainland} \quotes{cheap}\\ 
    Target sentence & \chinese{其他 電器 衫褲 好多 都 平 過 大陸}\\
    & \quotes{other} \quotes{electrical appliances} \quotes{clothes} \quotes{many} \quotes{all} \quotes{cheap} \quotes{than} \quotes{mainland}\\ 
    Our system & \chinese{其他 電器 服裝 等 有 好多 都 平 過 大陸}\\ 
    & \quotes{other} \quotes{electrical appliances} \quotes{clothes} \quotes{etc.} \quotes{have} \quotes{many} \quotes{all} \quotes{cheap} \quotes{than} \quotes{mainland}\\ 
    Baidu Fanyi & \chinese{其他o既 電器 衫 等 有 好多 都 比 大陸 平}\\ 
    & \quotes{other} \quotes{electrical appliances} \quotes{clothes} \quotes{etc.} \quotes{have} \quotes{many} \quotes{all} \quotes{compare to} \quotes{mainland} \quotes{cheap}\\ 
    \midrule
    
    \toprule
    Example 3 & (Kaifang Cidian) \\
    \midrule
    Source sentence & \chinese{躺 到 床 上 去，讓 我 檢查 一下}\\
    & \quotes{lay} \quotes{to} \quotes{bed} \quotes{on} [particle], \quotes{let} \quotes{me} \quotes{check} \quotes{a bit}\\ 
    Target sentence & \chinese{訓 上 張 床，等 我 檢查 下}\\
    & \quotes{lay} \quotes{on} [quantifier] \quotes{bed}, \quotes{let} \quotes{me} \quotes{check} \quotes{a bit}\\ 
    Our system & \chinese{攤 咗 上 床 ， 等 我 check 一 check }\\ 
    & \quotes{lay} \quotes{already} \quotes{on} \quotes{bed}, \quotes{let} \quotes{me} \quotes{check} \quotes{one} \quotes{check}\\ 
    Baidu Fanyi & \chinese{攤 到 床 上 去，等 我 檢查 一下}\\ 
    & \quotes{lay} \quotes{to} \quotes{bed} \quotes{on} [particle], \quotes{let} \quotes{me} \quotes{check} \quotes{a bit}\\ 
    \midrule
    
    \toprule
    Example 4 & (Kaifang Cidian) \\
    \midrule
    Source sentence & \chinese{這陣子 每天 晚上 都 加班，幹得 累 成 一 灘 泥}\\
    & \quotes{recently} \quotes{every day} \quotes{evening} \quotes{all} \quotes{work overtime}, \quotes{working like} \quotes{exhausted} \quotes{as} \quotes{a} [quantifier] \quotes{mud}\\ 
    Target sentence & \chinese{呢排 晚晚 加班，做到 一隻 屐 噉}\\
    & \quotes{recently} \quotes{every night} \quotes{work overtime}, \quotes{working like} \quotes{a} \quotes{clog} \quotes{so}\\ 
    Our system & \chinese{呢輪 晚晚 都 開o.t.， 做到 一 頭 泥 }\\ 
    & \quotes{recently} \quotes{every night} \quotes{all} \quotes{work overtime}, \quotes{working like} \quotes{a} [quantifier] \quotes{mud}\\ 
    Baidu Fanyi & \chinese{呢排 晚晚 都 開OT，搞到 [支力] 到 一 灘 泥}\\ 
    & \quotes{recently} \quotes{every night} \quotes{all} \quotes{work overtime}, \quotes{until} \quotes{exhausted} \quotes{as} \quotes{a} [quantifier] \quotes{mud}\\ 
    \midrule
  \bottomrule
\end{tabular}
\end{table*}

%%
%% The next two lines define the bibliography style to be used, and
%% the bibliography file.
\bibliographystyle{ACM-Reference-Format}
\bibliography{main}

\end{document}